\documentclass{article}

\usepackage{microtype}
\usepackage{graphicx}
\usepackage{booktabs} 
\usepackage{url}            
\usepackage{booktabs}       
\usepackage{amsfonts}       
\usepackage{amsmath}
\usepackage{nicefrac}       
\usepackage{xcolor}
\usepackage{textcomp}
\usepackage{subfig}
\usepackage{array,multirow}
\usepackage[nomessages]{fp}
\usepackage{relsize}
\usepackage{caption}
\newcommand{\mc}{\mathcal}
\newcommand{\mb}{\boldsymbol}
\newcommand{\mbb}{\mathbb}

\usepackage{amsthm}

\newcommand{\tbf}[1]{\textbf{#1}}
\newcommand{\figref}[1]{Fig.~\ref{#1}}

\usepackage{tikz}
\usetikzlibrary{positioning,fit,calc,decorations.pathreplacing,arrows,backgrounds,shapes}
\usetikzlibrary{automata,arrows}

\newcommand{\nonl}{\renewcommand{\nl}{\let\nl\oldnl}}
\DeclareMathSizes{10}{9}{8}{8}

\usepackage{hyperref}

\usepackage[accepted]{icml2020}

\icmltitlerunning{You say Normalizing Flows I see Bayesian Networks}

\begin{document}

\twocolumn[
\icmltitle{You say Normalizing Flows I see Bayesian Networks}



\icmlsetsymbol{equal}{*}

\begin{icmlauthorlist}
\icmlauthor{Antoine Wehenkel}{ULiege}
\icmlauthor{Gilles Louppe}{ULiege}
\end{icmlauthorlist}

\icmlaffiliation{ULiege}{University of Li{\`e}ge}

\icmlcorrespondingauthor{Antoine Wehenkel}{antoine.wehenkel@uliege.be}

\icmlkeywords{Machine Learning, ICML}

\vskip 0.3in
]



\printAffiliationsAndNotice{}  

\begin{abstract}
Normalizing flows have emerged as an important family of deep neural networks for modelling complex probability distributions. 
In this note, we revisit their coupling and autoregressive transformation layers as probabilistic graphical models and show that they reduce to Bayesian networks  with a pre-defined topology and a learnable density at each node.
From this new perspective, we provide three results.
First, we show that stacking multiple transformations in a normalizing flow relaxes independence assumptions and entangles the model distribution.
Second, we show that a fundamental leap of capacity emerges when the depth of affine flows exceeds 3 transformation layers. 
Third, we prove the non-universality of the affine normalizing flow, regardless of its depth. 

\end{abstract}

\section{Introduction}
Normalizing flows~\citep[NF, ][]{NF} have gained popularity in the recent years because of their unique ability to model complex data distributions while allowing both for sampling and exact density computation. 
This family of deep neural networks combines a base distribution with a series of invertible transformations while keeping track of the change of density that is caused by each transformation.


Probabilistic graphical models (PGMs) are well-established mathematical tools that combine graph and probability theory to ease the manipulation of joint distributions. They are commonly used to visualize and reason about the set of independencies in probabilistic models. 
Among PGMs, Bayesian networks~\citep[BN, ][]{Pearl-BN} offer a nice balance between readability and modeling capacity. Reading independencies stated by a BN is simple and can be performed graphically with the d-separation algorithm \citep{d-separation}. 

In this note, we revisit NFs as Bayesian networks.
We first briefly review the mathematical grounds of these two worlds. Then, for the first time in the literature, we show that the modeling assumptions behind coupling and autoregressive transformations can be perfectly expressed by distinct classes of BNs. 
From this insight, we show that stacking multiple transformation layers relaxes independencies and entangles the model distribution.
Then, we show that a fundamental change of regime emerges when the NF architecture includes 3 transformation steps or more.
Finally, we prove the non-universality of affine normalizing flows.


\section{Background}
\subsection{Normalizing flows}
A normalizing flow is defined as a sequence of invertible transformation steps $\mb{g}_k : \mathbb{R}^d \to \mathbb{R}^d$  ($k=1, ..., K$) that are composed together to create an expressive invertible mapping $\mb{g} = \mb{g}_1 \circ \dots \circ \mb{g}_K : \mathbb{R}^d \to \mathbb{R}^d$. 
This mapping can be used to perform density estimation, using $\mb{g}(\cdot ;\mb{\theta}): \mbb{R}^d \rightarrow \mbb{R}^d$ to map a sample $\mb{x} \in \mathbb{R}^d$ to a latent vector $\mb{z} \in \mbb{R}^d$ equipped with a density $p_{\mb{z}}(\mb{z})$.
The transformation $\mb{g}$ implicitly defines a density $p(\mb{x}; \mb{\theta})$ as given by the change of variables formula,
\begin{equation*}
    p(\mb{x}; \mb{\theta}) = p_{\mb{z}}(\mb{g}(\mb{x};\mb{\theta})) \left| \det  J_{\mb{g}(\mb{x};\mb{\theta})} \right|, \label{eq:NF_DE}
\end{equation*}
where $J_{\mb{g}(\mb{x};\mb{\theta})}$ is the Jacobian of $\mb{g}(\mb{x};\mb{\theta})$ with respect to $\mb x$.
The resulting model is trained by maximizing the likelihood of the data $\{\mb{x}^1, ..., \mb{x}^N\}$. NFs can also be used for data generation tasks while keeping track of the density of the generated samples such as to improve the latent distribution in variational auto-encoders \citep{NF}.
In the rest of this paper, we will not distinguish between $\mb{g}$ and $\mb{g}_k$ when the discussion will be focused on only one of these steps $\mb{g}_k$.

In general, steps $\mb{g}$ can take any form as long as they define a bijective map. Here, we focus on a sub-class of normalizing flows for which these steps can be mathematically described as 
\begin{equation*}
    \mb{g}(\mb{x}) = \begin{bmatrix}
g^1(x_{1}; \mb{c}^1(\mb{x})) & \hdots & g^d(x_{d}; \mb{c}^d(\mb{x}))
\end{bmatrix},\label{eq:gnf}
\end{equation*}
where the $\mb{c}^i$ are denoted as the \tbf{conditioners} and constrain the structure of the Jacobian of $\mb{g}$. The functions $g^i$, partially parameterized by their conditioner, must be invertible with respect to their input variable $x_i$. These are usually defined as affine or strictly monotonic functions, with the latter being the most general class of invertible scalar continuous functions. 
In this note, we mainly discuss affine normalizers that can be expressed as 
$g(x;m, s) = x\exp(s) + m$
where $m \in \mathbb{R}$ and $s \in \mathbb{R}$ are computed by the conditioner.

\subsection{Bayesian networks}

Bayesian networks allow for a compact and natural representation of probability distributions by exploiting conditional independence. More precisely, a BN is a directed acyclic graph (DAG) which structure encodes for the conditional independencies through the concept of d-separation~\citep{d-separation}. Equivalently, its skeleton supports an efficient factorization of the joint distribution. 

A BN is able to model a distribution $p$ if and only if it is an I-map with respect to $p$. That is, iff the set of independencies stated by the BN structure is a subset of the independencies that holds for $p$. Equivalently, a BN is a valid representation of a random vector $\mb{x}$ iff its density $p_{\mb{x}}(\mb{x})$ can be factorized by the BN structure as 
\begin{equation}
    p_{\mb{x}}(\mb{x}) = \prod^d_{i=1}p(x_i|\mathcal{P}_i),\label{eq:BN-fact}
\end{equation} 
where  $\mathcal{P}_i = \{j: A_{i,j} = 1 \}$ denotes the set of parents of the vertex $i$ and $A \in \{0, 1\}^{d\times d}$ is the adjacency matrix of the BN. As an example, \figref{fig:mono-step-flows-a} is a valid BN for any distribution over $\mb{x}$ because it does not state any independence, leading to a factorization that results in the chain rule.

\begin{figure}
\centering
\subfloat[]{\label{fig:mono-step-flows-a}
   \begin{tikzpicture}[
          node distance=.5cm and .35cm,
          mynode/.style={draw,circle,text width=.35cm,align=center},
          var_x/.style={draw, circle, text width=.35cm, align=center}
        ]
            \node[var_x] (x1) {$x_1$};
            \node[var_x, right=of x1] (x2) {$x_2$};
            \node[var_x, below=of x1] (x3) {$x_3$};
            \node[var_x, right=of x3] (x4) {$x_4$};
            \path (x1) edge[-latex] (x2);
            \path (x1) edge[-latex] (x3);
            \path (x1) edge[-latex] (x4);
            \path (x2) edge[-latex] (x3);
            \path (x2) edge[-latex] (x4);
            \path (x3) edge[-latex] (x4);
        \end{tikzpicture}
    }
\subfloat[]{\label{fig:mono-step-flows-b}
   \begin{tikzpicture}[
          node distance=.5cm and .35cm,
          mynode/.style={draw,circle,text width=.35cm,align=center},
          var_x/.style={draw, circle, text width=.35cm, align=center}
        ]
            \node[var_x] (x1) {$x_1$};
            \node[var_x, right=of x1] (x2) {$x_2$};
            \node[var_x, below=of x1] (x3) {$x_3$};
            \node[var_x, right=of x3] (x4) {$x_4$};
            \path (x1) edge[-latex] (x3);
            \path (x1) edge[-latex] (x4);
            \path (x2) edge[-latex] (x3);
            \path (x2) edge[-latex] (x4);
    \end{tikzpicture}
   }
   \subfloat[]{\label{fig:mono-step-flows-c}
        \begin{tikzpicture}[
          node distance=.5cm and .35cm,
          var_z/.style={draw,circle,text width=.35cm,align=center},
          var_x/.style={draw, circle, double, text width=.35cm, align=center}
        ]
            \node[var_x] (x1) {$x_1$};
            \node[var_x, right=of x1] (x2) {$x_2$};
            \node[var_x, below=of x1] (x3) {$x_3$};
            \node[var_x, right=of x3] (x4) {$x_4$};
            \node[var_z, left=of x1] (z1) {$z_1$};
            \node[var_z, right=of x2] (z2) {$z_2$};
            \node[var_z, left=of x3] (z3) {$z_3$};
            \node[var_z, right=of x4] (z4) {$z_4$};
            \path (x1) edge[-latex] (x3);
            \path (x1) edge[-latex] (x4);
            \path (x2) edge[-latex] (x3);
            \path (x2) edge[-latex] (x4);
            \foreach \x in {1,...,4}
                \path (z\x) edge (x\x);
        \end{tikzpicture}}
\caption{Bayesian networks for single-step normalizing flows on a vector $\mb{x} = [x_1, x_2, x_3, x_4]^T$. (\tbf{a}) BN for an autoregressive conditioner. (\tbf{b}) BN for a coupling conditioner. (\tbf{c}) Pseudo BN for a coupling conditioner, with the latent variables shown explicitly. Double circles stand for deterministic functions of the parents and non-directed edges stand for bijective relationships.} \label{fig:mono-step-flows}
    \end{figure}
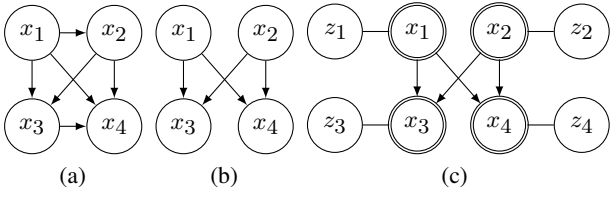
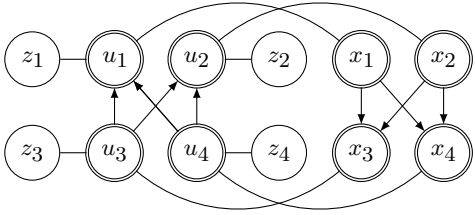
\begin{figure}
    \centering
        \begin{tikzpicture}[
          node distance=.5cm and .35cm,
          mynode/.style={draw,circle,text width=.35cm,align=center},
          var_x/.style={draw, circle, double, text width=.35cm, align=center}
        ]
            
            \node[var_x] (u1) {$u_1$};
            \node[var_x, right=of u1] (u2) {$u_2$};
            \node[var_x, below=of u1] (u3) {$u_3$};
            \node[var_x, right=of u3] (u4) {$u_4$};
            
            \node[mynode, left=of u1] (z1) {$z_1$};
            \node[mynode, right=of u2] (z2) {$z_2$};
            \node[mynode, left=of u3] (z3) {$z_3$};
            \node[mynode, right=of u4] (z4) {$z_4$};
            
            \node[var_x, right=of z2] (x1) {$x_1$};
            \node[var_x, right=of x1] (x2) {$x_2$};
            \node[var_x, right=of z4] (x3) {$x_3$};
            \node[var_x, right=of x3] (x4) {$x_4$};
            \path (x1) edge[-latex] (x3);
            \path (x1) edge[-latex] (x4);
            \path (x2) edge[-latex] (x3);
            \path (x2) edge[-latex] (x4);
            \foreach \x in {1,2}
                \path (u\x) edge[bend left=40] (x\x);
            
            \foreach \x in {3,4}
                \path (u\x) edge[bend left=-40] (x\x);
                
            \path (u4) edge[-latex] (u2);
            \path (u4) edge[-latex] (u1);
            \path (u3) edge[-latex] (u2);
            \path (u3) edge[-latex] (u1);
            \path (u4) edge[-latex] (u1);
            \foreach \x in {1, 2, 3,4}
                \path (z\x) edge (u\x);
        \end{tikzpicture}

    \caption{A Bayesian network equivalent to a 2-step normalizing flow based on coupling layers. Independence statements are relaxed by the second step.}
    \label{fig:two-steps-flows}
    \vspace{-1.em}
\end{figure}
    \vspace{-1.em}

\section{Normalizing flows as Bayesian networks}
\subsection{Autoregressive conditioners}
Autoregressive conditioners can be expressed as 
$$\mb{c}^i(\mb{x}) = \mb{h}^i\left(\begin{bmatrix} x_1 & \hdots & x_{i-1} \end{bmatrix}^T\right),$$ where $\mb{h}^i: \mathbb{R}^{i-1} \rightarrow \mathbb{R}^l$ are functions of the first $i-1$ components of $\mb{x}$ and whose output size depends on architectural choices.
These conditioners constrain the Jacobian of $\mb{g}$ to be lower triangular, making the computation of its determinant $\mc{O}(d)$. 
The multivariate density $p(\mb{x}; \mb{\theta})$ induced by $\mb{g}(\mb{x};\mb{\theta})$ and $p_{\mb{z}}(\mb{z})$ can be expressed as a product of $d$ univariate conditional densities, 
\begin{align}
    p(\mb{x}; \mb{\theta}) = p(x_1; \mb{\theta})\prod^{d}_{i=2}p(x_{i}|\mb{x}_{1:i-1}; \mb{\theta}). \label{eq:AF-fact}
\end{align}
When $p_{\mb{z}}(\mb{z})$ is a factored distribution $p_{\mb{z}}(\mb{z}) = \prod^{d}_{i=1}p(z_i)$, we identify that each component $z_i$ coupled with the corresponding function $g^i$ encodes for the conditional $p(x_{i}|\mb{x}_{1:i-1}; \mb{\theta})$. 
An explicit connection between BNs and autoregressive conditioners can be made if we define $\mathcal{P}_i = \{x_1, \hdots, x_{i-1}\}$ and compare \eqref{eq:AF-fact} with \eqref{eq:BN-fact}.  Therefore, and as illustrated in \figref{fig:mono-step-flows-a}, autoregressive conditioners can be seen as a way to model the conditional factors of a BN that does not state any independence. 
\subsection{Coupling conditioners}
Coupling conditioners~\cite{RealNVP} are another popular type of conditioners used in normalizing flows. The conditioners $\mb{c}^i$ made from coupling layers are defined as
\begin{align*}
    \mb{c}^i(\mb{x}) = 
    \begin{cases} 
    \underline{\mb{h}}^i \quad \text{if} \quad i < k\\
    \mb{h}^i(\mb{x}_{<k}) \quad \text{if} \quad i \geq k\\
    \end{cases}
\end{align*}
where the $\underline{\mb{h}}^i$ symbol define constant values. As for autoregressive conditioners, the Jacobian of $\mb{g}$ made of coupling layers is lower triangular. Assuming a factored latent distribution, the density associated with these conditioners can be written as follows:
\begin{align*}
    p(\mb{x}; \mb{\theta}) &= \prod^{k-1}_{i=1} p(x_i)\prod^{d}_{i=k} p(x_i|\mb{x}_{< k}),\\
    \text{where} \quad & p(x_i) = p(g^i(x_i; \underline{\mb{h}}^i))\frac{\partial g^i(x_i; \underline{\mb{h}}^i)}{\partial x_i} \\ \text{and} \quad & p(x_i|\mb{x}_{< k}) = p(g^i(x_i; \mb{h}^i(\mb{x}_{<k})) \frac{\partial g^i(x_i; \mb{h}^i(\mb{x}_{<k}))}{\partial x_i}.
\end{align*}
The factors define valid 1D conditional probability distributions because they can be seen as 1D changes of variables between $z_i$ and $x_i$. This factorization can be graphically expressed by a BN as shown in \figref{fig:mono-step-flows-b}. In addition, we can see \figref{fig:mono-step-flows-b} as the marginal BN of \figref{fig:mono-step-flows-c} which fully describes the stochastic process modeled by a NF that is made of a single transformation step and a coupling conditioner. 
In contrast to autoregressive conditioners, coupling layers are not by themselves universal density approximators, even when associated with very expressive normalizers $g^i$. 
Indeed, d-separation reveals independencies stated by this class of BN, such as the conditional independence between each pair in $\mathbf{x}_{\geq k}$ knowing $\mathbf{x}_{< k}$. These independence statements do not hold in general.

\subsection{Stacking transformation steps} \label{sec:multiple-step-flow-as-BN}

In practice, the invertible transformations discussed above are often stacked together in order to increase the representation capacity of the flow, with the popular good practice of permuting the vector components between two transformation steps.
The structural benefits of this stacking strategy can be explained from the perspective of the underlying BN.

First, a BN that explicitly includes latent variables is faithful as long as the sub-graph made only of those latent nodes is an I-map with respect to their distribution.
Normalizing flows composed of multiple transformation layers can therefore be viewed as single transformation flows whose latent distribution is itself recursively modeled by a normalizing flow.
As an example, \figref{fig:two-steps-flows} illustrates a NF made of two transformation steps with coupling conditioners. It can be observed that the latent vector $\mb{u}$ is itself a normalizing flow  whose distribution can be factored out by a class of BN. 

Second, from the BN associated to a NF, we observe that additional layers relax the independence assumptions defined by its conditioners.
The distribution modeled by the flow gets more entangled at each additional layer.
For example, \figref{fig:two-steps-flows} shows that for coupling layers, the additional steps relax the strong conditional independencies between $x_1$ and $x_2$ of the single transformation NF of \figref{fig:mono-step-flows-c}.
Indeed, we can observe from the figure that $x_1$ and $x_2$ have common ancestors ($z_3$ and $z_4$) whereas they are clearly assumed independent in \figref{fig:mono-step-flows-b}.

In general, we note that edges between two nodes in a BN do not model dependence, only the absence of edges does model independence. However, because some of the relationship between nodes are bijective, this implies that these nodes are strictly dependent on each other. We represent these relationships with undirected edges in the BN, as it can be seen in \figref{fig:two-steps-flows}.

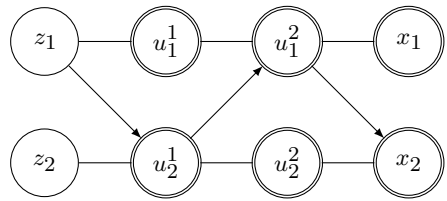
\begin{figure}
    \centering
        \begin{tikzpicture}[
            node distance=.7cm and .7cm,
            mynode/.style={draw,circle,text width=.35cm,align=center, minimum width=.9cm},
          var_x/.style={draw, circle, double, text width=.35cm, align=center, minimum width=.9cm}
        ]
            
            \node[mynode] (z1) {$z_1$};
            \node[mynode, below=of z1] (z2) {$z_2$};
            
            \node[var_x, right=of z1] (u11) {$u_1^1$};
            \node[var_x, right=of z2] (u12) {$u_2^1$};
            
            \node[var_x, right=of u11] (u21) {$u_1^2$};
            \node[var_x, right=of u12] (u22) {$u_2^2$};
            
            \node[var_x, right=of u21] (x1) {$x_1$};
            \node[var_x, right=of u22] (x2) {$x_2$};
            
            \path (z1) edge (u11);
            \path (z2) edge (u12);
            \path (z1) edge[-latex] (u12);
            
            \path (u11) edge (u21);
            \path (u12) edge (u22);
            \path (u12) edge[-latex] (u21);
            
            \path (u21) edge (x1);
            \path (u22) edge (x2);
            \path (u21) edge[-latex] (x2);
            
        \end{tikzpicture}
     \caption{The Bayesian network of a three-steps normalizing flow on vector $\mathbf{x} = [x_1, x_2]^T \in \mathbb{R}^4$. It can be observed that the distribution of the intermediate latent variables, and at the end of the vector $\mb{x}$, becomes more entangled at each additional transformation step. Considering the undirected edges as affine relationships, we see that while $u_1^1$ and $u_2^1$ are affine transformations of the latent $\mb{z}$, the vector $\mb{x}$ cannot be expressed as a linear function of the latent $\mb{z}$.}
     \label{fig:2d-three-steps-flows}
     \vspace{-1.5em}

\end{figure}

\section{Affine normalizing flows unlock their capacity with 3 transformation steps}

We now show how some of the limitations of affine normalizers can be relaxed by stacking multiple transformation steps. 
We also discuss why some limitations cannot be relaxed even with a large number of transformation steps. We intentionally put aside monotonic normalizers because they have already been proven to lead to universal density approximators when the conditioner is autoregressive~\cite{NAF}. 
We focus our discussion on a multivariate normal with an identity covariance matrix as base distribution $p_{\mb{z}}(\mb{z})$. 

We first observe from \figref{fig:mono-step-flows} that in a NF with a single transformation step at least one component of $\mb{x}$ is a function of only one latent variable.
If the normalizer is affine and the base distribution is normal, then this necessarily implies that the marginal distribution of this component is normal as well, which will very likely not lead to a good fit.
We easily see that adding steps relaxes this constraint.
A more interesting question to ask is what exactly the modeling capacity gain for each additional step of affine normalizer is. Shall we add steps to increase capacity or shall we increase the capacity of each step instead?
We first discuss a simple 2-dimensional case, which has the advantage of unifying the discussion for autoregressive and coupling conditioners, and then extend it to a more general setting.

Affine NFs made of a single transformation step induce strong constraints on the form of the density. In particular, these models implicitly assume that the data distribution can be factorized as a product of conditional normal distributions. 
These assumptions are rexaled when accumulating steps in the NF.
As an example, \figref{fig:2d-three-steps-flows} shows the equivalent BN of a 2D NF composed of 3 steps. This flow is mathematically described with the following set of equations:
\begin{align*}
\vspace{-5em}
\scriptstyle
    u_1^1 &:= z_1  & u_2^1 &:= \exp(s^1_2(z_1))z_2 + m_2^1(z_1) \\
    u_2^2 &:= u_2^1 & u_1^2 &:= \exp(s^2_1\left(u_2^1\right))z_1 + m^2_1\left(u_2^1\right)  \\
    x_1 &:=  u_1^2 & 
    x_2 &:= \exp(s_2^3\left(u_1^2\right))u^2_2 + m^3_2\left( u_1^2 \right)
\end{align*} 
From these equations, we see that after one step the latent variables $u^1_1$ and $u_2^1$ are respectively normal and conditionally normal. 
This is relaxed with the second step, where the latent variable $u^2_1$ is a non-linear function of two random variables distributed normally (by assumption on the distribution of $z_1$ and $z_2$). 
However, $u_2
^2$ is a stochastic affine transformation of a normal random variable. 
In addition, we observe that the expression of $u^2_1$ is strictly more expressive than the expression of $u^2_2$. 
Finally, $x_1$ and $x_2$ are non-linear functions of both latent variables $z_1$ and $z_2$. 
Assuming that the functions $s^i_j$ and $m^i_j$ are universal approximators, we argue that the stochastic process that generates $x_1$ and the one that generates $x_2$ are as expressive as each other. 
Indeed, by making the functions arbitrarily complex the transformation for $x_1$ could be made arbitrarily close to the transformations for $x_2$ and vice versa. 
This is true because both transformations can be seen as an affine transformation of a normal random variables whose scaling and offset factors are non-linear arbitrarily expressive transformations of all the latent variables. Because of this equilibrium between the two expressions, additional steps do not improve the modeling capacity of the flow.
The same observations can be made empirically as illustrated in \figref{fig:2d-three-steps-flows} for 2-dimensional toy problems. A clear leap of capacity occurs from 2-step to 3-step NFs, while having 4 steps or more does not result in any noticeable improvement when $s^i_j$ and $m^i_j$ already have enough capacity.

For $d > 2$, autoregressive and coupling conditioners do not correspond to the same set of equations or BN. However, if the order of the vector is reversed between two transformation steps, the discussion generalizes to any value of $d$ for both conditioners. Indeed, in both cases each component of the intermediate latent vectors can be seen as having a set of conditioning variables and a set of independent variables. At each successive step the indices of the non-conditioning variables are exchanged with the conditioning ones and thus any output vector's component can be expressed either as a component of the vector form of $x_1$ or of $x_2$.

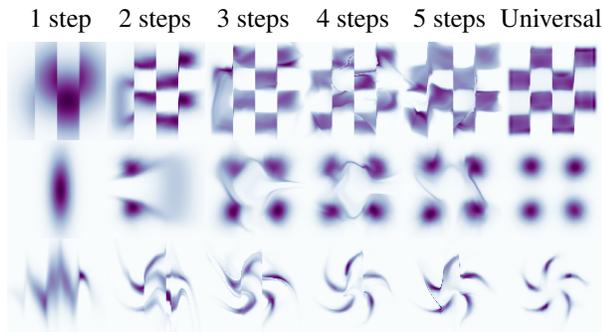
\begin{figure}
    \def\layersep{2.5cm}
    \centering
    \begin{tikzpicture}[shorten >=1pt,->,draw=black!50, node distance=1.25cm, scale=0.45]
    \tikzstyle{every pin edge}=[<-,shorten <=1pt]
    \tikzstyle{neuron}=[circle,fill=black!25,minimum size=7pt,inner sep=0pt]
    \tikzstyle{input neuron}=[neuron, fill=black!50];
    \tikzstyle{output neuron}=[neuron, fill=black!50];
    \tikzstyle{hidden neuron}=[neuron, fill=black!50];
    \tikzstyle{annot} = [text width=4em, text centered]
    \tikzset{edge/.style = {->,-latex}}

    \node[inner sep=0pt] (graphical) at (0,0)
    {\includegraphics[width=.46\textwidth]{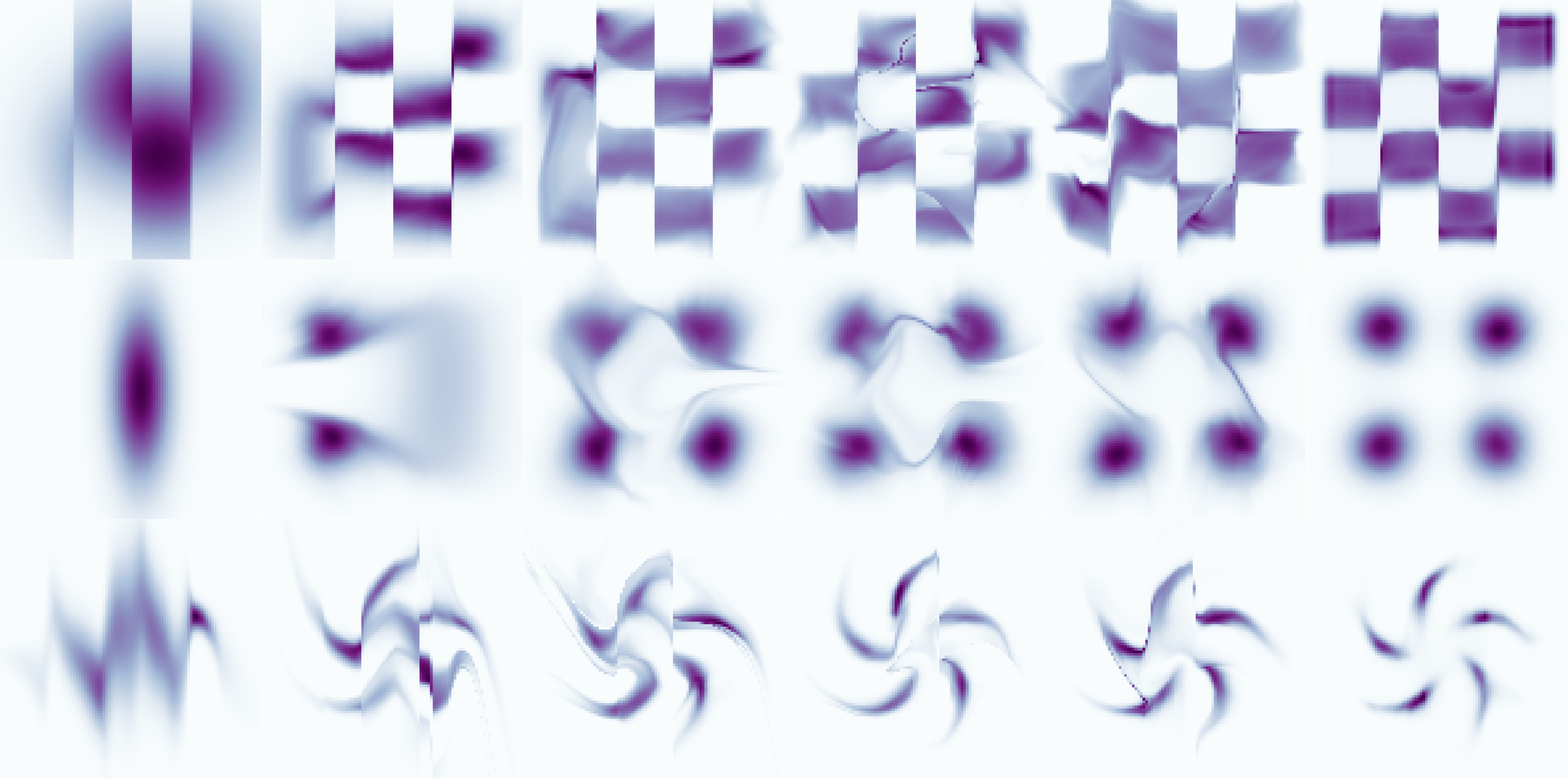}};
    \node[] () at (-7.2, 5) {1 step};
    \node[] () at (-4.4, 5) {2 steps};
    \node[] () at (-1.5, 5) {3 steps};
    \node[] () at (1.45, 5) {4 steps};
    \node[] () at (4.3, 5) {5 steps};
    \node[] () at (7.3, 5.08) {Universal};
\end{tikzpicture}
\hspace{-1.5em}
    \caption{Evolution of an affine normalizing flow's capacity as the number of steps increases. For comparison, the density learned by a universal density approximator is shown on the last column.}
    \label{fig:BN-steps}
    \vspace{-1.em}

\end{figure}
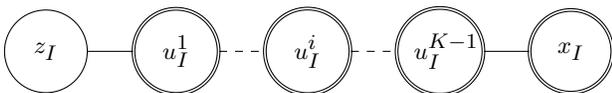
\begin{figure}
    \centering
    \begin{tikzpicture}[
            node distance=.6cm and .6cm,
            mynode/.style={draw,circle,text width=.69cm,align=center, minimum width=1.1cm},
          var_x/.style={draw, circle, double, text width=.69cm, align=center, minimum width=1.1cm}
        ]
            \node[mynode] (zI) {$z_I$};
            \node[var_x, right=of zI] (u1I) {$u^1_I$};
            \node[var_x, right=of u1I] (uiI) {$u^i_I$};
            \node[var_x, right=of uiI] (uKI) {$u^{\scriptstyle K-1}_I$};
            \node[var_x, right=of uKI] (xI) {$x_I$};
            \path (zI) edge (u1I);
            \path (u1I) edge[dashed] (uiI);
            \path (uiI) edge[dashed] (uKI);
            \path (uKI) edge (xI);
        \end{tikzpicture}
    \hspace{-1.5em}

    \caption{The equivalent BN of a component with a unique latent variable as ancestor.}
    \label{fig:BN-x-I}
    \vspace{-2.2em}
\end{figure}

\section{Affine normalizing flows are not universal density approximators}

We argue that affine normalizers do not lead to universal density approximators in general, even for an infinite number of steps. In the following, we assume again that the latent variables are distributed according to a normal distribution with a unit covariance matrix.

To prove the non-universality of affine normalizing flows, one only needs to provide a counter-example.
Let us consider the simple setup in which one component $x_I$ of the random vector $\mb x$ is independent from all the other components.
Let us also assume that $x_I$ is distributed under a non-normal distribution.
We can then consider two cases.
First, $x_I$ has only one component of the latent vector $\mb z$ as an ancestor. This implies that the equivalent BN would be as in \figref{fig:BN-x-I}, hence that $x_I$ is a linear function of this ancestor and is therefore normally distributed.
Else, $x_I$ has $n$ components of the latent vector as ancestors.
However, this second case would imply that at least one undirected edge is removed from the original BN considered in Section~\ref{sec:multiple-step-flow-as-BN}.
This cannot happen since it would deadly hurt the bijectivity of the flow.

Besides proving the non-universality of affine NFs, this discussion provides the important insight that when affine normalizers must transform non-linearly some latent variables they introduce dependence in the model of the distribution. In some sense, this means that the additional disorder required to model this non-normal component is performed at the cost of some loss in entropy caused by mutual information between the random vector components. 
\section{Summary}

In this preliminary work, we have revisited normalizing flows from the perspective of Bayesian networks.
We have shown that stacking multiple transformations in a normalizing flow relaxes independence assumptions and entangles the model distribution.
Then, we have shown that affine normalizing flows benefit from having at least 3 transformation layers. 
Finally, we demonstrated that they remain non-universal density approximators regardless of their depths.

We hope these results will give practitioners more intuition in the design of normalizing flows.
We also believe that this work may lead to further research. 
First, unifying Bayesian networks and normalizing flows could be pushed one step further with conditioners that are specifically designed to model Bayesian networks. 
Second, the study could be extended for other type of normalizing flows such as non-autoregressive monotonic flows. 
Finally, we believe this study may spark research at the intersection of structural equation modeling, causal networks and normalizing flows.
\newpage
\subsection*{Acknowledgments}
The authors would like to thank Matthia Sabatelli, Johann Brehmer  and Louis Wehenkel for proofreading the manuscript. 
Antoine Wehenkel is a research fellow of the F.R.S.-FNRS (Belgium) and acknowledges its financial support.
Gilles Louppe is recipient of the ULi{\`e}ge - NRB Chair on Big data and is thankful for the support of NRB.
\bibliographystyle{abbrv}
\bibliography{innf}
\end{document}